\documentclass[11pt,a4paper]{article}
\usepackage[hyperref]{emnlp2018}
\usepackage{times}
\usepackage{latexsym}

\usepackage{amsmath}
\usepackage{amsfonts}
\usepackage{amssymb}
\usepackage{graphicx}
\usepackage{caption}
\usepackage{subcaption}
\usepackage{array}
\usepackage{booktabs}
\usepackage{xcolor}
\usepackage{dsfont}

\usepackage{url}

\aclfinalcopy 

\newcommand\enc{^{\rm{enc}}}
\newcommand\dec{^{\rm{dec}}}
\newcommand\tmp{^{\rm{tmp}}}
\newcommand\z{^{\rm{z}}}
\newcommand\intra{^{\rm{int}}}
\newcommand\emb{^{\rm{emb}}}
\newcommand\hid{^{\rm{hid}}}
\newcommand\real{{\mathbb{R}}}
\newcommand\gen{^{\rm{gen}}}
\newcommand\cop{^{\rm{cp}}}
\newcommand\softmax{{\rm{softmax}}}
\newcommand\sigmoid{{\rm{sigmoid}}}
\newcommand\ml{^{\rm{ml}}}
\newcommand\pg{^{\rm{pg}}}
\newcommand\sample{^{\rm{sam}}}
\newcommand\greedy{^{\rm{gre}}}
\newcommand\lm{^{\rm{lm}}}
\newcommand\fuse{^{\rm{fuse}}}
\newcommand\generated{x\gen}
\newcommand\sourcedoc{x^{\rm{src}}}
\newcommand\groundtruth{x^{\rm{gt}}}
\newcommand\rouge{^{\rm{rou}}}
\newcommand\novel{^{\rm{nov}}}
\newcommand{\norm}[1]{\left\lVert#1\right\rVert}
\newcommand{\ngram}[1]{{\rm{ng}}\left(#1\right)}

\title{Improving Abstraction in Text Summarization}

\author{Wojciech Kry\'sci\'nski
  \thanks{\hspace*{0.6em}Work performed while at Salesforce Research.}\\ 
  KTH Royal Institute of Technology \\
  {\tt wkr@kth.se} \\\And
  Romain Paulus \\
  Salesforce Research \\
  {\tt rpaulus@salesforce.com} \\\AND
  Caiming Xiong \\
  Salesforce Research \\
  {\tt cxiong@salesforce.com} \\\And
  Richard Socher \\
  Salesforce Research \\
  {\tt rsocher@salesforce.com} \\}

\date{}

\begin{document}
\maketitle

\begin{abstract}
Abstractive text summarization aims to shorten long text documents into a human readable form that contains the most important facts from the original document.
However, the level of actual abstraction as measured by novel phrases that do not appear in the source document remains low in existing approaches.
We propose two techniques to improve the level of abstraction of generated summaries.
First, we decompose the decoder into a contextual network that retrieves relevant parts of the source document, and a pretrained language model that incorporates prior knowledge about language generation.
Second, we propose a novelty metric that is optimized directly through policy learning to encourage the generation of novel phrases.
Our model achieves results comparable to state-of-the-art models, as determined by ROUGE scores and human evaluations, while achieving a significantly higher level of abstraction as measured by $n$-gram overlap with the source document.
\end{abstract}

\section{Introduction}

\begin{table*}
\centering
\resizebox{\linewidth}{!}{
\begin{tabular}{l}
\toprule
{\bf Article} \\
\midrule
(cnn) to allay possible concerns, boston prosecutors released video friday of the shooting of a police officer last month that\\
resulted in the killing of the gunman. the officer wounded, john moynihan, is white. angelo west, the gunman shot to death \\
by officers, was black. after the shooting, community leaders in the predominantly african-american neighborhood of (...) \\
\midrule
{\bf Human-written summary} \\
\midrule
boston police officer john moynihan is \colorbox{yellow}{released from the hospital.} video shows that the man later shot dead by police in \\
boston opened fire first. moynihan was shot \colorbox{yellow}{in the face} during a traffic stop. \\
\midrule
{\bf Generated summary \citep{See:17}} \\
\midrule
\colorbox{yellow}{boston prosecutors released video friday of the shooting of a police officer last month. the gunman shot to death by officers}, \\
\colorbox{yellow}{was black}. \colorbox{yellow}{one said the officers were forced to return fire.} \colorbox{yellow}{he was placed in a medically induced coma at a boston hospital.} \\
\midrule
{\bf Generated summary \citep{Liu:18}} \\
\midrule
\colorbox{yellow}{boston prosecutors released video} \colorbox{yellow}{of the shooting of a police officer last month}. \colorbox{yellow}{the shooting occurred} \colorbox{yellow}{in the wake of the} \\
\colorbox{yellow}{boston marathon bombing.} \colorbox{yellow}{the video shows} \colorbox{yellow}{west sprang out and fired a shot with a pistol} at \colorbox{yellow}{officer's face.} \\
\midrule
{\bf Our summary (ML+RL ROUGE+Novel, with LM)} \\ 
\midrule
new: boston police release video of shooting of officer , john moynihan. new: angelo \colorbox{yellow}{west had several prior gun convictions}, \\
police say. boston police officer john \colorbox{yellow}{moynihan, 34, survived with a bullet wound}\colorbox{yellow}{. he was} in a \colorbox{yellow}{medically induced coma at} \\
\colorbox{yellow}{a boston hospital}, a police officer says. \\

\bottomrule
\end{tabular}
}
\caption{Summaries generated by different models for the same CNN/Daily Mail article. The highlighted spans indicate phrases of 3 tokens or more that are copied word-by-word from the original article.}
\label{tab:example-summaries}
\end{table*}

Text summarization concerns the task of compressing a long sequence of text into a more concise form.
The two most common approaches to summarization are \textit{extractive} \citep{Dorr:03, Nallapati:17}, where the model extracts salient parts of the source document, and \textit{abstractive} \citep{Paulus:17, See:17}, where the model not only extracts but also concisely paraphrases the important parts of the document via generation.
We focus on developing a summarization model that produces an increased level of abstraction.
That is, the model produces concise summaries without only copying long passages from the source document.

A high quality summary is shorter than the original document,
conveys only the most important and no extraneous information,
and is semantically and syntactically correct.
Because it is difficult to gauge the correctness of the summary, evaluation metrics for summarization models use word overlap with the ground-truth summary in the form of ROUGE \citep{Lin:04} scores.
However, word overlap metrics do not capture the \textit{abstractive} nature of high quality human-written summaries: the use of paraphrases with words that do not necessarily appear in the source document.

The state-of-the-art abstractive text summarization models have high word overlap performance, however they tend to copy long passages of the source document directly into the summary, thereby producing summaries that are not abstractive~\citep{See:17}.

We propose two general extensions to summarization models that improve the level of abstraction of the summary while preserving word overlap with the ground-truth summary.
Our first contribution decouples the extraction and generation responsibilities of the decoder by factoring it into a contextual network and a language model.
The contextual network has the sole responsibility of extracting and compacting the source document whereas the language model is responsible for the generation of concise paraphrases.
Our second contribution is a mixed objective that jointly optimizes the $n$-gram overlap with the ground-truth summary while encouraging abstraction.
This is done by combining maximum likelihood estimation with policy gradient.
We reward the policy with the ROUGE metric, which measures word overlap with the ground-truth summary, as well as a novel abstraction reward that encourages the generation of words not in the source document.

We demonstrate the effectiveness of our contributions on a encoder-decoder summarization model.
Our model obtains state-of-the-art ROUGE-L scores, and ROUGE-1 and ROUGE-2 performance comparable to state-of-the-art methods on the CNN/DailyMail dataset.
Moreover, we significantly outperform all previous abstractive approaches in our abstraction metrics. Table \ref{tab:example-summaries} shows a comparison of summaries generated by our model and previous abstractive models, showing less copying and more abstraction in our model.

\begin{figure*}
\centering
\includegraphics[width=1.0\linewidth]{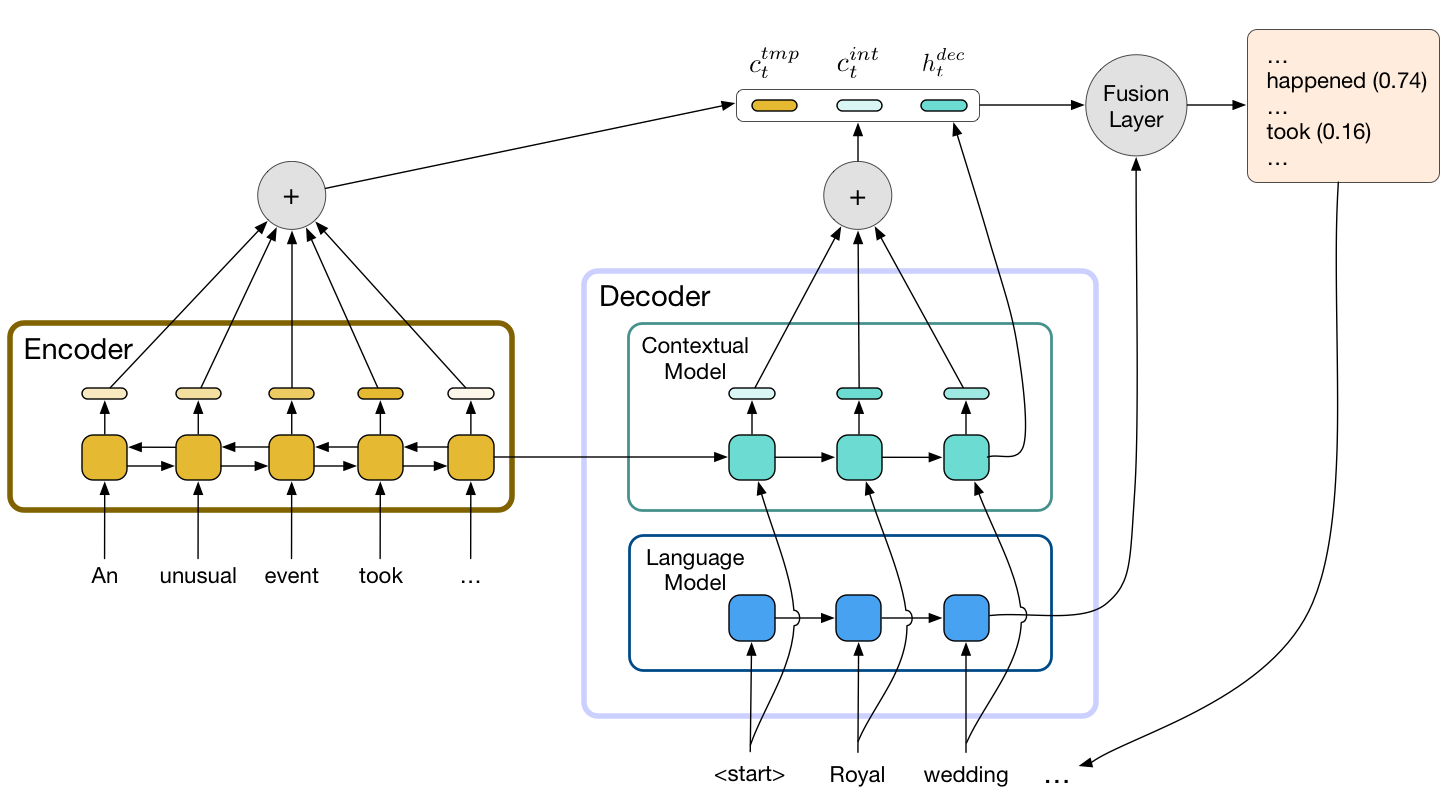}
\caption{The network architecture with the decoder factorized into separate contextual and language models. The reference vector, composed of context vectors $c^{tmp}_t$, $c^{int}_t$, and the hidden state of the contextual model $h^{dec}_t$, is fused with the hidden state of the language model and then used to compute the distribution over the output vocabulary.}
\label{fig:system-diagram}
\end{figure*}

\section{Model}

\subsection{Base Model and Training Objective}
The base model follows the encoder-decoder architecture with temporal attention and intra-attention proposed by \citet{Paulus:17}.
Let $E \in \real^{n \times d\emb}$ denote the matrix of $d\emb$ dimensional word embeddings of the $n$ words in the source document.
The encoding of the source document $h\enc$ is computed via a bidirectional LSTM \citep{Hochreiter:97} whose output has dimension $d\hid$.

\begin{equation}
h\enc = {\rm{BiLSTM}}\left(E\right) \in \real^{n \times d\hid}
\label{eq:enc-hidden}
\end{equation}

The decoder uses temporal attention over the encoded sequence that penalizes input tokens that previously had high attention scores.
Let $h\dec_t$ denote the decoder state at time $t$.
The temporal attention context at time $t$, $c\tmp_t$, is computed as

\begin{eqnarray}
s\tmp_{ti} &=&
    \left( h\dec_t \right)^\intercal
    W\tmp 
    h\enc_i
    \in \real \\
q\tmp_{ti} &=&
    \frac 
        {\exp(s\tmp_{ti})}
        {\sum^{t-1}_{j=1}\exp(s\tmp_{ji})} 
    \in \real \\
\alpha\tmp_{ti} &=&
        \frac{q\tmp_{ti}}{\sum^n_{j=1} q\tmp_{tj}} 
    \in \real \\
c\tmp_{t} &=&
    \sum^n_{i=1}
    \alpha\tmp_{ti}
    h\enc_i
    \in \real^{d\hid}
\end{eqnarray}

where we set $q\tmp_{ti}$ to $\exp(s\tmp_{ti})$ for $t = 1$.

The decoder also attends to its previous states via intra-attention over the decoded sequence.
The intra-attention context at time $t$, $c\intra_t$, is computed as

\begin{eqnarray}
s\intra_{ti} &=&
    \left( h^{dec}_t \right)^\intercal
    W\intra
    h\dec_i
    \in \real \\
c\intra_{t} &=&
    \sum^{t-1}_{i=1}
    \left(
        \frac{s\intra_{ti}}{\sum^n_{j=1} s\intra_{tj}} 
    \right) h\dec_i
        \label{eq:attn-intra}
    \in \real^{d\hid}
\end{eqnarray}

The decoder generates tokens by interpolating between selecting words from the source document via a pointer network as well as selecting words from a fixed output vocabulary.
Let $z_t$ denote the ground-truth label as to whether the $t$th output word should be generated by the selecting from the output vocabulary as opposed to from the source document.
We compute $p(z_t)$, the probability that the decoder generates from the output vocabulary, as

\begin{eqnarray}
r_t &= [h\dec_t; c\tmp_t; c\intra_t] \in \real^{3 d\hid} \label{eq:rt} \\
p(z_t) &= \sigmoid( W\z r_t + b\z ) \in \real
\end{eqnarray}

The probability of selecting the word $y_t$ from a fixed vocabulary at time step $t$ is defined as

\begin{eqnarray}
p\gen (y_t) = \softmax \left( W\gen r_t + b\gen \right)
\label{eq:p_y_t}
\end{eqnarray}

We set $p\cop (y_t)$, the probability of copying the word $y_t$ from the source document, to the temporal attention distribution $\alpha\tmp_t$.
The joint probability of using the generator and generating the word $y_t$ at time step $t$, $p(z_t, y_t)$, is then

\begin{eqnarray}
p (z_t, y_t) = p(y_t \mid z_t) p(z_t) \label{eq:joint}
\end{eqnarray}

the likelihood of which is

\begin{equation}
\begin{split}
&\log p (z_t, y_t) = \log p(y_t \mid z_t) + \log p(z_t) \\
&= z_t \log p\gen(y_t) + (1-z_t) \log p\cop(y_t) \\
&\text{\quad} + z_t \log p(z_t) + \left(1-z_t\right) \log \left(1-p\left(z_t\right)\right) \\
&= z_t \left( \log p\gen(y_t) + \log p(z_t) \right) \\
&\text{\quad} + (1-z_t) \left( \log p\cop(y_t) + \log \left( 1-p\left(z_t\right) \right) \right)
\end{split}
\end{equation}

The objective function combines maximum likelihood estimation with policy learning.
Let $m$ denote the length of the ground-truth summary, 
The maximum likelihood loss $L\ml$ is computed as

\begin{eqnarray}
L\ml = - \sum^m_{t=1} \log p (z_t, y_t) \label{eq:xe-loss}
\end{eqnarray}

Policy learning uses ROUGE-L as its reward function and a self-critical baseline using the greedy decoding policy \citep{Rennie:16}.
Let $y\sample$ denote the summary obtained by sampling from the current policy $p$, $y\greedy$ and $z\greedy$ the summary and generator choice obtained by greedily choosing from $p(z_t, y_t)$, $R(y)$ the ROUGE-L score of the summary $y$, and $\Theta$ the model parameters.
The policy learning loss is

\begin{eqnarray}
\hat{R} &=& R \left( y\sample \right) - R \left( y\greedy \right)\\
L\pg &=& - \mathbb{E}_{\substack{z\sample~\sim p(z),\\y\sample~\sim p(y\vert z)}} [ \hat{R} ] \label{eq:policygradient}
\end{eqnarray}

where we use greedy predictions by the model according to eq.~\eqref{eq:xe-loss} as a baseline for variance reduction.
The policy gradient, as per~\citet{Schulman:15}, is

\begin{eqnarray}
\nabla_\Theta L\pg \approx - \hat{R} \sum_{t=1}^m \nabla_\Theta \log p \left( z\sample_t, y\sample_t \right)
\end{eqnarray}

The final loss is a mixture between the maximum likelihood loss and the policy learning loss, weighted by a hyperparameter $\gamma$.

\begin{equation}
L = (1 - \gamma) L\ml + \gamma L\pg
\label{eq:objective}
\end{equation}

\subsection{Language Model Fusion}

The decoder is an essential component of the base model.
Given the source document and the previously generated summary tokens, the decoder both extracts relevant parts of the source document through the pointer network as well as composes paraphrases from the fixed vocabulary.
We decouple these two responsibilities by augmenting the decoder with an external language model.
The language model assumes responsibility of generating from the fixed vocabulary, and allows the decoder to focus on attention and extraction.
This decomposition has the added benefit of easily incorporating external knowledge about fluency or domain specific styles via pre-training the language model on a large scale text corpora.

The architecture of our language model is based on~\citet{Merity:17}.
We use a 3-layer unidirectional LSTM with weight-dropped LSTM units.

Let $e_t$ denote the embedding of the word generated during time step $t$.
The hidden state of the language model at the $l$-th layer is

\begin{eqnarray}
h\lm_{l,t} = {\rm{LSTM}}\lm_3 \left(e_{t-1}, h\lm_{l,t-1} \right)
\end{eqnarray} 

At each time step $t$, we combine the hidden state of the last language model LSTM layer, $h\lm_{3,t}$, with $r_t$ defined in eq.~\eqref{eq:rt} in a fashion similar to~\citet{Sriram:17}.
Let $\odot$ denote element-wise multiplication.
We use a gating function whose output $g_t$ filters the content of the language model hidden state.
\begin{eqnarray}
f_t &=& \sigmoid \left(W\lm [r_t; h\lm_{3,t}] + b\lm \right) \\
g_t &=& W\fuse ([r_t; g_t \odot h\lm_{3,t}]) + b\fuse\\
h\fuse_t &=& {\rm{ReLU}} \left( g_t \right)
\end{eqnarray}

We then replace the output distribution of the language model $p\gen \left( y_t \right)$ in eq. \ref{eq:p_y_t} with
\begin{eqnarray}
p\gen \left( y_t \right) = \softmax \left( W\gen h\fuse_t + b\gen \right)
\end{eqnarray}

\subsection{Abstractive Reward}
In order to produce an abstractive summary, the model cannot exclusively copy from the source document.
In particular, the model needs to parse large chunks of the source document and create concise summaries using phrases not in the source document.
To encourage this behavior, we propose a novelty metric that promotes the generation of novel words.

We define a \textit{novel} phrase in the summary as one that is not in the source document.
Let $\ngram{x, n}$ denote the function that computes the set of unique $n$-grams in a document $x$, $\generated$ the generated summary, $\sourcedoc$ the source document, and $\norm{s}$ the number of words in $s$.
The unnormalized novelty metric $N$ is defined as the fraction of unique $n$-grams in the summary that are novel.

\begin{eqnarray}
N \left(\generated, n\right) = \frac{\norm{\ngram{\generated, n} - \ngram{\sourcedoc, n}}}{\norm{\ngram{\generated, n}}}
\end{eqnarray}

To prevent the model for receiving high novelty rewards by outputting very short summaries, we normalize the metric by the length ratio of the generated and ground-truth summaries.
Let $\groundtruth$ denote the ground-truth summary.
We define the novelty metric as

\begin{eqnarray}
R\novel \left(\generated, n\right) =
    N \left(\generated, n\right)
    \frac{\norm{\generated}}{\norm{\groundtruth}}
\end{eqnarray}

We incorporate the novelty metric as a reward into the policy gradient objective in eq.~\eqref{eq:policygradient}, alongside the original ROUGE-L metric.
In doing so, we encourage the model to generate summaries that both overlap with human written ground-truth summaries as well as incorporate novel words not in the source document:

\begin{eqnarray}
R\left(y\right) =
    \lambda\rouge R\rouge \left(y\sample\right)
    +
    \lambda\novel R\novel \left(y\sample\right)
\end{eqnarray}

where $\lambda\rouge$ and $\lambda\novel$ are hyperparameters that control the weighting of each reward.

\section{Experiments}

\subsection{Datasets}
We train our model on the CNN/Daily Mail dataset \citep{Hermann:15,Nallapati:16}.
Previous works on abstractive summarization either use an anonymized version of this dataset or the original article and summary texts.
Due to these different formats, it is difficult to compare the overall ROUGE scores and performance between each version.
In order to compare against previous results, we train and evaluate on both versions of this dataset.
For the anonymized version, we follow the pre-processing steps described in \citet{Nallapati:16}, and the pre-processing steps of \citet{See:17} for the the full-text version.

\begin{table*}[t]
\centering
\resizebox{\linewidth}{!}{%
\begin{tabular}{l|rrr|rrrr}
\toprule
{\bf Model} & {\bf R-1} & {\bf R-2} & {\bf R-L} & {\bf NN-1} & {\bf NN-2} & {\bf NN-3} & {\bf NN-4} \\
\midrule
\multicolumn{8}{c}{\textit{anonymized}}\\
\midrule

\textit{Ground-truth summaries} & - & - & - & \textit{14.40} & \textit{52.07} & \textit{71.63} & \textit{80.84}\\
ML+RL, intra-attn \citep{Paulus:17} & 39.87 & \textbf{15.82} & 36.9 & 1.04 & 10.86 & 21.53 & 29.27\\
\midrule

ML+RL ROUGE+Novel, with LM (ours) & \textbf{40.02} & 15.53 & \textbf{37.44} & \textbf{3.54} & \textbf{21.91} & \textbf{37.48} & \textbf{47.13}\\
\midrule
\multicolumn{8}{c}{\textit{full-text}}\\
\midrule
\textit{Ground-truth summaries} & - & - & - & \textit{13.55} & \textit{49.97} & \textit{70.32} & \textit{80.02} \\ 
Pointer-gen + coverage \citep{See:17} & 39.53 & 17.28 & 36.38 &  0.07 &  2.24 & 6.03 & 9.72\\
SumGAN \citep{Liu:18} & 39.92 & 17.65 & 36.71 & 0.22 & 3.15 & 7.68 & 11.84\\
RSal \citep{Pasunuru:18} & 40.36 & 17.97 & 37.00 & - & 2.37 & 6.00 & 9.50\\
RSal+Ent RL \citep{Pasunuru:18} & \textbf{40.43} & \textbf{18.00} & 37.10 & - & - & - & -\\
\midrule
ML+RL ROUGE+Novel, with LM (ours) & 40.19 & 17.38 & \textbf{37.52} & \textbf{3.25} & \textbf{17.21} & \textbf{30.46} & \textbf{39.47} \\
\bottomrule
\end{tabular}
}
\caption{Comparison of ROUGE (R-) and novel $n$-gram (NN-) test results for our model and other abstractive summarization models on the CNN/Daily Mail dataset.}
\label{tab:cnndm-results}
\end{table*}

We use named entities and the source document to supervise the model regarding when to use the pointer and when to use the generator (e.g. $z_t$ in eq.~\eqref{eq:xe-loss}.
Namely, during training, we teach the model to point from the source document if the word in the ground-truth summary is a named entity, an out-of-vocabulary word, or a numerical value that is in the source document.
We obtain the list of named entities from \citet{Hermann:15}.

\subsection{Language Models}
\label{sub:language-models}

For each dataset version, we train a language model consisting of a 400-dimensional word embedding layer and a 3-layer LSTM with each layer having a hidden size of 800 dimensions, except the last input layer which has an output size of 400.
The final decoding layer shares weights with the embedding layer \citep{Inan:16,Press:16}.
We also use DropConnect~\citep{Wan:13} in the hidden-to-hidden connections, as well as the non-monotonically triggered asynchronous gradient descent optimizer from~\citet{Merity:17}.

We train this language model on the CNN/Daily Mail ground-truth summaries only, following the same training, validation, and test splits as our main experiments. 

\subsection{Training details}

The two LSTMs of our bidirectional encoder are 200-dimensional, and out decoder LSTM is 400-dimensional. We restrict the input vocabulary for the embedding matrix to 150,000 tokens, and the output decoding layer to 50,000 tokens. We limit the size of input articles to the first 400 tokens, and the summaries to 100 tokens. We use scheduled sampling \citep{Bengio:15} with a probability of 0.25 when calculating the maximum-likelihood training loss. We also set $n=3$ when computing our novelty reward $R^{nov}(x_{gen}, n)$. For our final training loss using reinforcement learning, we set $\gamma=0.9984$, $\lambda^{rou}=0.9$, and $\lambda^{nov}=0.1$.
Finally, we use the trigram repetition avoidance heuristic defined by \citet{Paulus:17} during beam search decoding to ensure that the model does not output twice the same trigram in a given summary, reducing the amount of repetitions.

\subsection{Novelty baseline}
\label{sub:novelty-baseline}

We also create a novelty baseline by taking the outputs of our base model, without RL training and without the language model, and inserting random words not present in the article after each summary token with a probability $r=0.0005$. 
This baseline will intuitively have a higher percentage of novel $n$-grams than our base model outputs while being very similar to these original outputs, hence keeping the ROUGE score difference relatively small.

\section{Results}

\subsection{Quantitative analysis}

We obtain a validation and test perplexity of 65.80 and 66.61 respectively on the anonymized dataset, and 81.13 and 82.98 on the full-text dataset with the language models described in Section~\ref{sub:language-models}.

The ROUGE scores and novelty scores of our final summarization model on both versions of the CNN/Daily Mail dataset are shown in Table \ref{tab:cnndm-results}.
We report the ROUGE-1, ROUGE-2, and ROUGE-L F-scores as well as the percentage of novel $n$-grams, marked NN-$n$, in the generated summaries, with $n$ from 1 to 4. Results are omitted in cases where they have not been made available by previous authors. We also include the novel $n$-gram scores for the ground-truth summaries as a comparison to indicate the level of abstraction of human written summaries.

Even though our model outputs significantly fewer novel $n$-grams than human written summaries, it has a much higher percentage of novel $n$-grams than all the previous abstractive approaches. It also achieves state-of-the-art ROUGE-L performance on both dataset versions, and obtains ROUGE-1 and ROUGE-2 scores close to state-of-the-art results.

\subsection{Ablation study}
\label{sub:ablation-study}

\begin{table*}
\centering
\begin{tabular}{l|rrr|rrrr}
\toprule
{\bf Model} & {\bf R-1} & {\bf R-2} & {\bf R-L} & {\bf NN-1} & {\bf NN-2} & {\bf NN-3} & {\bf NN-4} \\
\midrule
ML & 39.21 & 15.47 & 36.27 & 2.47 & 14.1 & 25.35 & 33.46\\ 
ML with nov. baseline, $r=0.0005$ & 38.62 & 15.06 & 35.75 & 3.12 & 14.96 & 26.45 & 34.76\\ 
ML with LM & 39.43 & 15.68 & 36.45 & 3.36 & 15.25 & 26.06 & 33.57\\ 
\midrule
ML+RL ROUGE & 41.02 & 16.62 & 38.13 & 2.2 & 12.88 & 24.16 & 32.5\\ 
ML+RL ROUGE, with LM & \textbf{41.06} & \textbf{16.84} & 38.01 & 2.06 & 10.9 & 19.78 & 26.33\\ 
\midrule
ML+RL ROUGE+Novel & 40.61 & 15.84 & 38.06& 3.19 & \textbf{22.79} & \textbf{39.9} & \textbf{50.61}\\ 
ML+RL ROUGE+Novel, with LM & 40.72 & 15.95 & \textbf{38.14} & \textbf{3.49} & 21.89 & 37.31 & 46.85\\ 
\bottomrule
\end{tabular}
\caption{Ablation study on the validation set of the anonymized CNN/Daily Mail dataset.}
\label{tab:cnndm-ablation-study}
\end{table*}

In order to evaluate the relative impact of each of our individual contributions, we run ablation studies comparing our model ablations against each other and against the novelty baseline.
The results of these different models on the validation set of the anonymized CNN/Daily Mail dataset are shown in Table \ref{tab:cnndm-ablation-study}.
Results show that our base model trained with the maximum-likelihood loss only and using the language model in the decoder (ML, with LM) has higher ROUGE scores, novel unigrams, and novel bigrams scores than our base model without the language model (ML). ML with LM also beats the novelty baseline for these metrics.
When training these models with reinforcement learning using the ROUGE reward (ML+RL ROUGE and ML+RL ROUGE with LM), the model with language model obtains higher ROUGE-1 and ROUGE-2 scores. However, it also loses its novel unigrams and bigrams advantage.
Finally, using the mixed ROUGE and novelty rewards (ML+RL ROUGE+Novel) produces both higher ROUGE scores and more novel unigrams with the language model than without it.
This indicates that the combination of the language model in the decoder and the novelty reward during training makes our model produce more novel unigrams while maintaining high ROUGE scores.

\subsection{ROUGE vs novelty trade-off}

\begin{figure*}[t]
\centering
\begin{subfigure}{0.49\linewidth}
  \includegraphics[width=1.0\linewidth]{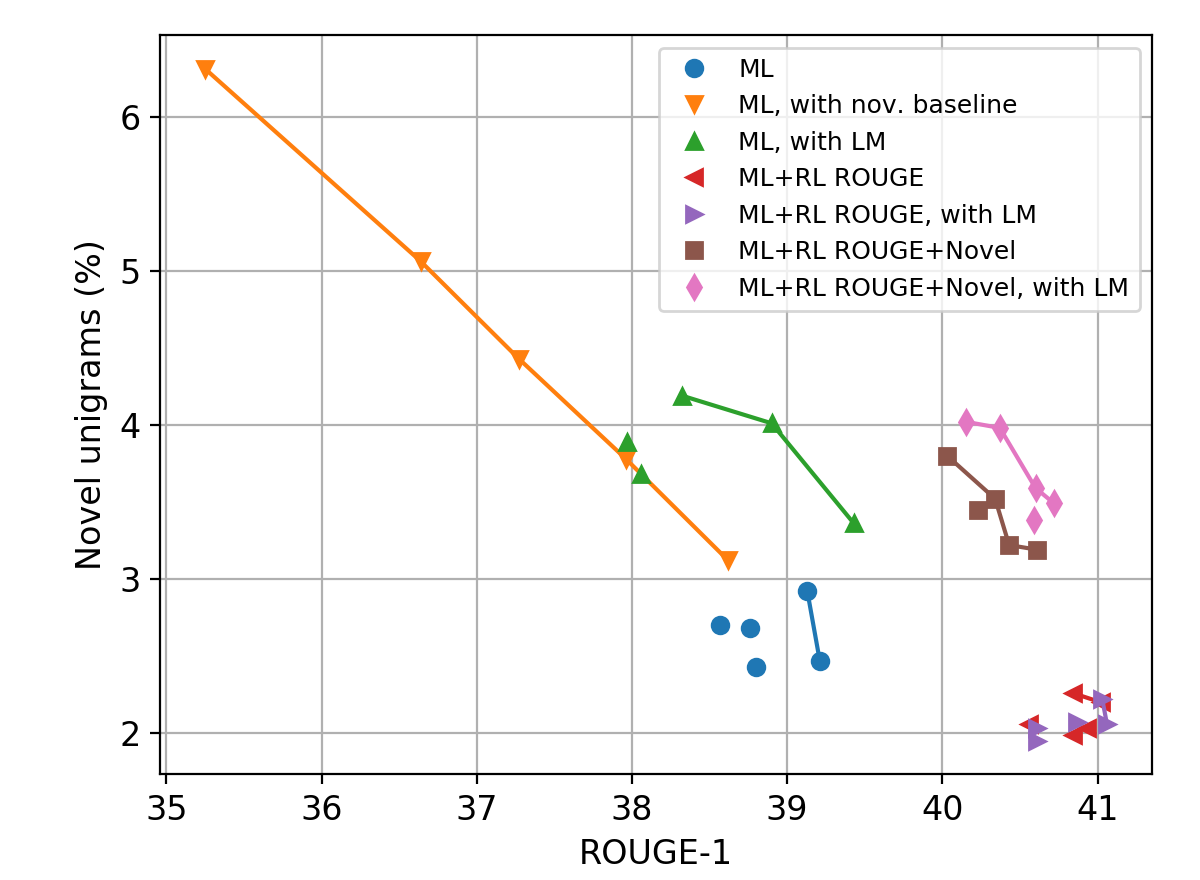}
\end{subfigure}
\begin{subfigure}{0.49\linewidth}
  \includegraphics[width=1.0\linewidth]{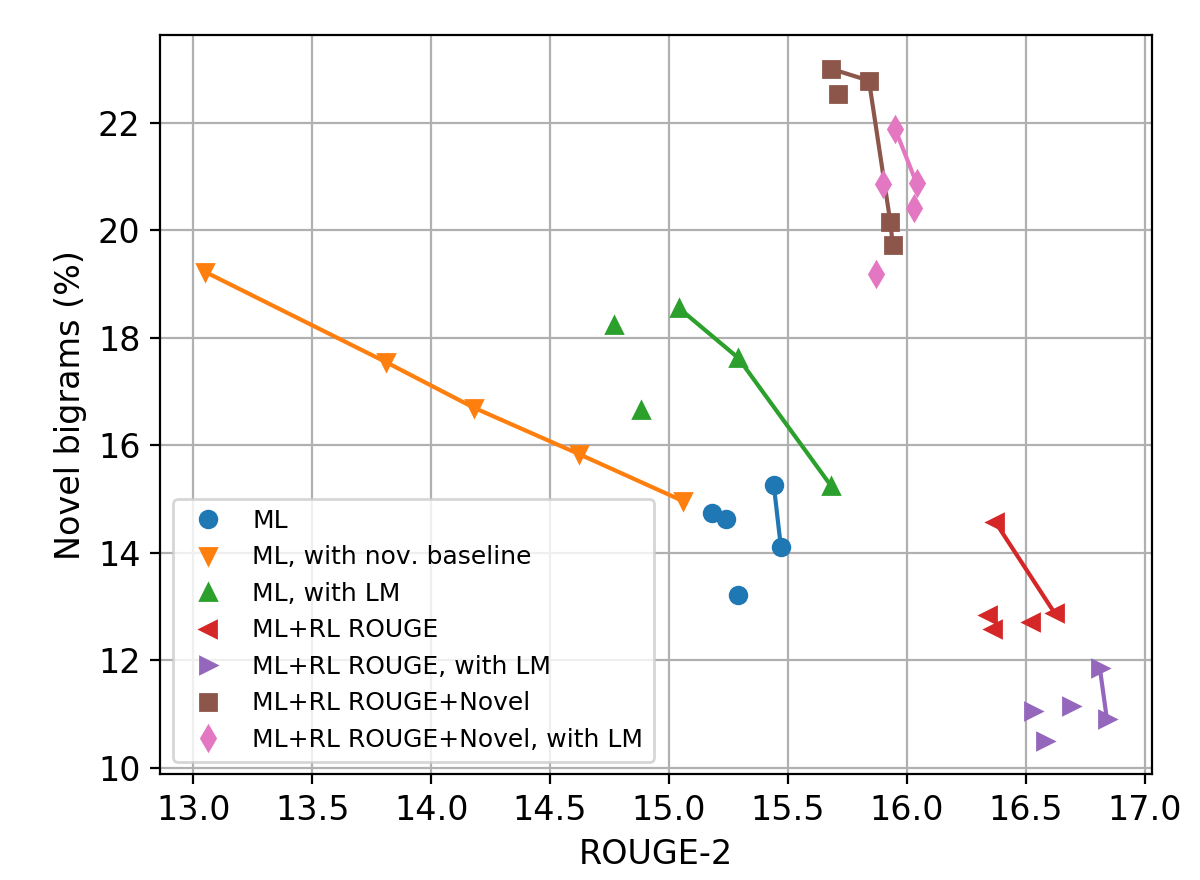}
\end{subfigure}
\caption{ROUGE and novel $n$-grams results on the anonymized validation set for different runs of each model type. Lines indicates the Pareto frontier for each model type.}
\label{fig:rouge-abstraction-tradeoff}
\end{figure*}

\begin{table*}[t]
\centering
\begin{tabular}{l|rr}
\toprule
{\bf Model} & {\bf Readability} & {\bf Relevance} \\
\midrule
Pointer-gen + coverage \citep{See:17} & \textbf{6.76} $\pm$ 0.17 & \textbf{6.73} $\pm$ 0.17\\
SumGAN \citep{Liu:18} & 6.79 $\pm$ 0.16 & 6.74 $\pm$ 0.17\\
\midrule
ML+RL ROUGE+Novel, with LM & 6.35 $\pm$ 0.19 & 6.63 $\pm$ 0.18\\
\bottomrule
\end{tabular}
\caption{Mean and confidence interval at 95\% of human evaluation scores on the full-text test outputs. Individual summaries are rated from 1 to 10, a higher score indicating higher quality, for readability and relevance separately.}
\label{tab:human-evaluation}
\end{table*}

In order to understand the correlation between ROUGE and novel $n$-gram scores across different architectures, and to find the model type that gives the best trade-off between each of these metrics, we plot the ROUGE-1 and novel unigram scores for the five best iterations of each model type on the anonymized dataset, as well as the ROUGE-2 and novel bigram scores on a separate plot.
We also include the novelty baseline described in Section \ref{sub:ablation-study} for values of $r$ between 0.005 and 0.035.
For each model type, we indicate the Pareto frontier by a line plot \citep{Ben:1980}, illustrating which models of a given type give the best combination of ROUGE and novelty scores.
These plots are shown in Figure \ref{fig:rouge-abstraction-tradeoff}.

These plots show that there exist an inverse correlation between ROUGE and novelty scores in all model types, illustrating the challenge of choosing a model that performs well in both. Given that, our final model (ML+RL ROUGE+Novel, with LM) provides the best trade-off of ROUGE-1 scores compared to novel unigrams, indicated by the higher Pareto frontier in the first plot. Similarly, our final model gives one of the best trade-offs of ROUGE-2 scores to novel bigrams, even though the same model without LM produces more novel bigrams with a lower ROUGE-2 score.

\subsection{Qualitative evaluation}

In order to ensure the quality of our model outputs, we ask 5 human evaluators to rate 100 randomly selected full-text test summaries, giving them two scores from 1 to 10 respectively for readability and relevance given the original article.
We also include the full-text test outputs from ~\citet{See:17} and ~\citet{Liu:18} for comparison.
Evaluators are shown different summaries corresponding to the same article side by side without being told which models have generated them.
The mean score and confidence interval at 95\% for each model and each evaluation criterion are reported in Table~\ref{tab:human-evaluation}. 
These results show that our model matches the relevance score of ~\citet{See:17} and ~\citet{Liu:18}, but is slightly inferior to them in terms of readability.

\section{Related work}

\paragraph{Text summarization.}
Existing summarization approaches are usually either extractive or abstractive. In extractive summarization, the model selects passages from the input document and combines them to form a shorter summary, sometimes with a post-processing step to ensure final coherence of the output ~\citep{Neto:02,Dorr:03,filippova2013,colmenares2015,Nallapati:17}.
While extractive models are usually robust and produce coherent summaries, they cannot create concise summaries that paraphrase the source document using new phrases.

Abstractive summarization allows the model to paraphrase the source document and create concise summaries with phrases not in the source document.
The state-of-the-art abstractive summarization models are based on sequence-to-sequence models with attention ~\citep{Bahdanau:14}.
Extensions to this model include a self-attention mechanism \citep{Paulus:17} and an article coverage vector \citep{See:17} to prevent repeated phrases in the output summary. Different training procedures have also been used improve the ROUGE score \citep{Paulus:17} or textual entailment \citep{Pasunuru:18} with reinforcement learning; as well as generative adversarial networks to generate more natural summaries \citep{Liu:18}.

Several datasets have been used to train and evaluate summarization models.
The Gigaword \citep{Graff:03} and some DUC datasets \citep{Over:07} have been used for headline generation models \citep{Rush:15,Nallapati:16}, where the generated summary is shorter than 75 characters. However, generating longer summaries is a more challenging task, especially for abstractive models. \citet{Nallapati:16} have proposed using the CNN/Daily Mail dataset \citep{Hermann:15} to train models for generating longer, multi-sentence summaries up to 100 words. The New York Times dataset \citep{Sandhaus:08} has also been used as a benchmark for the generation of long summaries \citep{Durrett:16,Paulus:17}.

\paragraph{Training strategies for sequential models.}
The common approach to training models for sequence generation is maximum likelihood estimation with teacher forcing.
At each time step, the model is given the previous ground-truth output and predicts the current output.
The sequence objective is the accumulation of cross entropy losses from each time step.

Despite its popularity, this approach for sequence generation is suboptimal due to exposure bias~\citep{Huszar:15} and loss-evaluation mismatch \citep{Wiseman:16}. 
\citet{Goyal:16} propose one way to reduce exposure bias by explicitly forcing the hidden representations of the model to be similar during training and inference.
\citet{Bengio:15} and \citet{Wiseman:16} propose an alternate method that exposes the network to the test dynamics during training.
Reinforcement learning methods \citep{Sutton:98}, such as policy learning \citep{Sutton:99}, mitigate the mismatch between the optimization objective and the evaluation metrics by directly optimizing evaluation metrics.
This approach has led to consistent improvements in domains such as image captioning \citep{Zhang:17} and abstractive text summarization \citep{Paulus:17}. 

A recent approach to training sequential models utilizes generative adversarial networks to improving the human perceived quality of generated outputs \citep{Fedus:18,Guimaraes:17,Liu:18}.
Such models use an additional discriminator network that distinguishes between natural and generated output to guide the generative model towards outputs akin to human-written text.

\section{Conclusions}

We introduced a new abstractive summarization model which uses an external language model in the decoder, as well as a new reinforcement learning reward to encourage summary abstraction. Experiments on the CNN/Daily Mail dataset show that our model generates summaries that are much more abstractive that previous approaches, while maintaining high ROUGE scores close to or above the state of the art.
Future work could be done on closing the gap to match human levels of abstraction, which is still very far ahead from our model in terms of novel $n$-grams. Including mechanisms to promote paraphrase generation in the summary generator could be an interesting direction.

\bibliography{emnlp2018}

\begin{thebibliography}{36}
\expandafter\ifx\csname natexlab\endcsname\relax\def\natexlab#1{#1}\fi

\bibitem[{Bahdanau et~al.(2015)Bahdanau, Cho, and Bengio}]{Bahdanau:14}
Dzmitry Bahdanau, Kyunghyun Cho, and Yoshua Bengio. 2015.
\newblock Neural machine translation by jointly learning to align and
  translate.
\newblock In \emph{ICLR}.

\bibitem[{Ben-Tal(1980)}]{Ben:1980}
Aharon Ben-Tal. 1980.
\newblock Characterization of pareto and lexicographic optimal solutions.
\newblock In \emph{Multiple Criteria Decision Making Theory and Application},
  pages 1--11. Springer.

\bibitem[{Bengio et~al.(2015)Bengio, Vinyals, Jaitly, and Shazeer}]{Bengio:15}
Samy Bengio, Oriol Vinyals, Navdeep Jaitly, and Noam Shazeer. 2015.
\newblock Scheduled sampling for sequence prediction with recurrent neural
  networks.
\newblock In \emph{NIPS}.

\bibitem[{Colmenares et~al.(2015)Colmenares, Litvak, Mantrach, and
  Silvestri}]{colmenares2015}
Carlos~A Colmenares, Marina Litvak, Amin Mantrach, and Fabrizio Silvestri.
  2015.
\newblock Heads: Headline generation as sequence prediction using an abstract
  feature-rich space.
\newblock In \emph{HLT-NAACL}, pages 133--142.

\bibitem[{Dorr et~al.(2003)Dorr, Zajic, and Schwartz}]{Dorr:03}
Bonnie Dorr, David Zajic, and Richard Schwartz. 2003.
\newblock Hedge trimmer: A parse-and-trim approach to headline generation.
\newblock In \emph{HLT-NAACL}.

\bibitem[{Durrett et~al.(2016)Durrett, Berg-Kirkpatrick, and
  Klein}]{Durrett:16}
Greg Durrett, Taylor Berg-Kirkpatrick, and Dan Klein. 2016.
\newblock Learning-based single-document summarization with compression and
  anaphoricity constraints.
\newblock In \emph{ACL}.

\bibitem[{Fedus et~al.(2018)Fedus, Goodfellow, and Dai}]{Fedus:18}
William Fedus, Ian~J. Goodfellow, and Andrew~M. Dai. 2018.
\newblock Maskgan: Better text generation via filling in the
  {\_}{\_}{\_}{\_}{\_}{\_}.
\newblock In \emph{ICLR}.

\bibitem[{Filippova and Altun(2013)}]{filippova2013}
Katja Filippova and Yasemin Altun. 2013.
\newblock Overcoming the lack of parallel data in sentence compression.
\newblock In \emph{Proceedings of EMNLP}, pages 1481--1491. Citeseer.

\bibitem[{Goyal et~al.(2016)Goyal, Lamb, Zhang, Zhang, Courville, and
  Bengio}]{Goyal:16}
Anirudh Goyal, Alex Lamb, Ying Zhang, Saizheng Zhang, Aaron~C. Courville, and
  Yoshua Bengio. 2016.
\newblock Professor forcing: {A} new algorithm for training recurrent networks.
\newblock In \emph{NIPS}.

\bibitem[{Graff and Cieri(2003)}]{Graff:03}
David Graff and C~Cieri. 2003.
\newblock English gigaword, linguistic data consortium.

\bibitem[{Guimaraes et~al.(2017)Guimaraes, Sanchez{-}Lengeling, Farias, and
  Aspuru{-}Guzik}]{Guimaraes:17}
Gabriel~Lima Guimaraes, Benjamin Sanchez{-}Lengeling, Pedro Luis~Cunha Farias,
  and Al{\'{a}}n Aspuru{-}Guzik. 2017.
\newblock Objective-reinforced generative adversarial networks {(ORGAN)} for
  sequence generation models.
\newblock \emph{CoRR}, abs/1705.10843.

\bibitem[{Hermann et~al.(2015)Hermann, Kocisky, Grefenstette, Espeholt, Kay,
  Suleyman, and Blunsom}]{Hermann:15}
Karl~Moritz Hermann, Tomas Kocisky, Edward Grefenstette, Lasse Espeholt, Will
  Kay, Mustafa Suleyman, and Phil Blunsom. 2015.
\newblock Teaching machines to read and comprehend.
\newblock In \emph{NIPS}.

\bibitem[{Hochreiter and Schmidhuber(1997)}]{Hochreiter:97}
Sepp Hochreiter and J{\"{u}}rgen Schmidhuber. 1997.
\newblock Long short-term memory.
\newblock \emph{Neural Computation}, 9(8):1735--1780.

\bibitem[{Huszar(2015)}]{Huszar:15}
Ferenc Huszar. 2015.
\newblock How (not) to train your generative model: Scheduled sampling,
  likelihood, adversary?
\newblock \emph{CoRR}, abs/1511.05101.

\bibitem[{Inan et~al.(2017)Inan, Khosravi, and Socher}]{Inan:16}
Hakan Inan, Khashayar Khosravi, and Richard Socher. 2017.
\newblock Tying word vectors and word classifiers: A loss framework for
  language modeling.
\newblock In \emph{ICLR}.

\bibitem[{Lin(2004)}]{Lin:04}
Chin-Yew Lin. 2004.
\newblock Rouge: A package for automatic evaluation of summaries.
\newblock In \emph{Proc. ACL workshop on Text Summarization Branches Out},
  page~10.

\bibitem[{Liu et~al.(2018)Liu, Lu, Yang, Qu, Zhu, and Li}]{Liu:18}
Linqing Liu, Yao Lu, Min Yang, Qiang Qu, Jia Zhu, and Hongyan Li. 2018.
\newblock Generative adversarial network for abstractive text summarization.
\newblock In \emph{AAAI}.

\bibitem[{Merity et~al.(2018)Merity, Keskar, and Socher}]{Merity:17}
Stephen Merity, Nitish~Shirish Keskar, and Richard Socher. 2018.
\newblock Regularizing and optimizing lstm language models.
\newblock In \emph{ICLR}.

\bibitem[{Nallapati et~al.(2017)Nallapati, Zhai, and Zhou}]{Nallapati:17}
Ramesh Nallapati, Feifei Zhai, and Bowen Zhou. 2017.
\newblock Summarunner: {A} recurrent neural network based sequence model for
  extractive summarization of documents.
\newblock In \emph{AAAI}.

\bibitem[{Nallapati et~al.(2016)Nallapati, Zhou, G{\"u}l{\c c}ehre, Xiang
  et~al.}]{Nallapati:16}
Ramesh Nallapati, Bowen Zhou, {\c C}a\u{g}lar G{\"u}l{\c c}ehre, Bing Xiang,
  et~al. 2016.
\newblock Abstractive text summarization using sequence-to-sequence rnns and
  beyond.
\newblock \emph{Proceedings of SIGNLL Conference on Computational Natural
  Language Learning}.

\bibitem[{Neto et~al.(2002)Neto, Freitas, and Kaestner}]{Neto:02}
Joel~Larocca Neto, Alex~A Freitas, and Celso~AA Kaestner. 2002.
\newblock Automatic text summarization using a machine learning approach.
\newblock In \emph{Brazilian Symposium on Artificial Intelligence}, pages
  205--215. Springer.

\bibitem[{Over et~al.(2007)Over, Dang, and Harman}]{Over:07}
Paul Over, Hoa Dang, and Donna Harman. 2007.
\newblock Duc in context.
\newblock \emph{Inf. Process. Manage.}, 43(6):1506--1520.

\bibitem[{Pasunuru and Bansal(2018)}]{Pasunuru:18}
Ramakanth Pasunuru and Mohit Bansal. 2018.
\newblock Multi-reward reinforced summarization with saliency and entailment.
\newblock \emph{CoRR}, abs/1804.06451.

\bibitem[{Paulus et~al.(2017)Paulus, Xiong, and Socher}]{Paulus:17}
Romain Paulus, Caiming Xiong, and Richard Socher. 2017.
\newblock A deep reinforced model for abstractive summarization.
\newblock In \emph{ICLR}.

\bibitem[{Press and Wolf(2016)}]{Press:16}
Ofir Press and Lior Wolf. 2016.
\newblock Using the output embedding to improve language models.
\newblock \emph{arXiv preprint arXiv:1608.05859}.

\bibitem[{Rennie et~al.(2016)Rennie, Marcheret, Mroueh, Ross, and
  Goel}]{Rennie:16}
Steven~J. Rennie, Etienne Marcheret, Youssef Mroueh, Jarret Ross, and Vaibhava
  Goel. 2016.
\newblock Self-critical sequence training for image captioning.
\newblock \emph{CoRR}, abs/1612.00563.

\bibitem[{Rush et~al.(2015)Rush, Chopra, and Weston}]{Rush:15}
Alexander~M Rush, Sumit Chopra, and Jason Weston. 2015.
\newblock A neural attention model for abstractive sentence summarization.
\newblock \emph{Proceedings of EMNLP}.

\bibitem[{Sandhaus(2008)}]{Sandhaus:08}
Evan Sandhaus. 2008.
\newblock The new york times annotated corpus.
\newblock \emph{Linguistic Data Consortium, Philadelphia}, 6(12):e26752.

\bibitem[{Schulman et~al.(2015)Schulman, Heess, Weber, and
  Abbeel}]{Schulman:15}
John Schulman, Nicolas Heess, Theophane Weber, and Pieter Abbeel. 2015.
\newblock Gradient estimation using stochastic computation graphs.
\newblock In \emph{NIPS}.

\bibitem[{See et~al.(2017)See, Liu, and Manning}]{See:17}
Abigail See, Peter~J. Liu, and Christopher~D. Manning. 2017.
\newblock Get to the point: Summarization with pointer-generator networks.
\newblock In \emph{ACL}.

\bibitem[{Sriram et~al.(2017)Sriram, Jun, Satheesh, and Coates}]{Sriram:17}
Anuroop Sriram, Heewoo Jun, Sanjeev Satheesh, and Adam Coates. 2017.
\newblock Cold fusion: Training seq2seq models together with language models.
\newblock \emph{CoRR}, abs/1708.06426.

\bibitem[{Sutton and Barto(1998)}]{Sutton:98}
Richard~S. Sutton and Andrew~G. Barto. 1998.
\newblock \emph{Reinforcement learning - an introduction}.
\newblock Adaptive computation and machine learning. {MIT} Press.

\bibitem[{Sutton et~al.(1999)Sutton, McAllester, Singh, and
  Mansour}]{Sutton:99}
Richard~S. Sutton, David~A. McAllester, Satinder~P. Singh, and Yishay Mansour.
  1999.
\newblock Policy gradient methods for reinforcement learning with function
  approximation.
\newblock In \emph{NIPS}.

\bibitem[{Wan et~al.(2013)Wan, Zeiler, Zhang, Le~Cun, and Fergus}]{Wan:13}
Li~Wan, Matthew Zeiler, Sixin Zhang, Yann Le~Cun, and Rob Fergus. 2013.
\newblock Regularization of neural networks using dropconnect.
\newblock In \emph{ICML}.

\bibitem[{Wiseman and Rush(2016)}]{Wiseman:16}
Sam Wiseman and Alexander~M. Rush. 2016.
\newblock Sequence-to-sequence learning as beam-search optimization.
\newblock In \emph{EMNLP}.

\bibitem[{Zhang et~al.(2017)Zhang, Sung, Liu, Xiang, Gong, Yang, and
  Hospedales}]{Zhang:17}
Li~Zhang, Flood Sung, Feng Liu, Tao Xiang, Shaogang Gong, Yongxin Yang, and
  Timothy~M. Hospedales. 2017.
\newblock Actor-critic sequence training for image captioning.
\newblock \emph{CoRR}, abs/1706.09601.

\end{thebibliography}
\bibliographystyle{acl_natbib_nourl}
\end{document}